\documentclass{article}

\usepackage[preprint,nonatbib]{nips_2018}

\usepackage[utf8]{inputenc} 
\usepackage[T1]{fontenc}    
\usepackage{hyperref}       
\usepackage{url}            
\usepackage{booktabs}       
\usepackage{amsfonts}       
\usepackage{nicefrac}       
\usepackage{microtype}      
\usepackage{listings}
\usepackage[utf8]{inputenc}
\usepackage{graphicx}
\usepackage{wrapfig}
\usepackage{amsmath,amsfonts,amssymb,amsthm}

\usepackage{color}
\usepackage{soul}  

\usepackage[ruled,vlined]{algorithm2e}
\graphicspath{{images/}}

\begin{document}

\title{Empowering Knowledge Distillation via Open Set Recognition for Robust 3D Point Cloud Classification}
\author{
Ayush Bhardwaj$^{*\dagger}$,
Sakshee Pimpale$^{\diamond\dagger}$,
Saurabh Kumar$^{\diamond\dagger}$,
Biplab Banerjee$^{\ddagger}$,
\\ [8pt]
$^{*}$Department of Physics, IIT Bombay \\
$^{\diamond}$Department of Electrical Engineering, IIT Bombay\\
$^{\ddagger}$Centre for Machine Intelligence and Data Science, IIT Bombay\\
}

\maketitle
\begin{abstract}
  Real-world scenarios pose several challenges to deep learning based computer vision techniques despite their tremendous success in research.
Deeper models provide better performance, but are challenging to deploy and knowledge distillation allows us to train smaller models with minimal loss in performance.
The model also has to deal with open set samples from classes outside the ones it was trained on and should be able to identify them as unknown samples while classifying the known ones correctly.
Finally, most existing image recognition research focuses only on using two-dimensional snapshots of the real world three-dimensional objects.
In this work, we aim to bridge these three research fields, which have been developed independently until now, despite being deeply interrelated.
We propose a joint Knowledge Distillation and Open Set recognition training methodology for three-dimensional object recognition.
We demonstrate the effectiveness of the proposed method via various experiments on how it allows us to obtain a much smaller model, which takes a minimal hit in performance while being capable of open set recognition for 3D point cloud data.
\end{abstract}

\section{Introduction}
Deep neural networks are currently the state of the art models for most computer vision tasks. 
Their adoption has significantly accelerated primarily due to the availability of cheaper computation power, massive training datasets, and powerful development frameworks.
Along with this, an increase in the number of parameters of these models has been shown to be directly correlated with their performance. 
In particular, deeper models with more parameters usually provide better performance than shallower models.
Due to this, there are significant research efforts to build and train larger and deeper models to push the performance frontier of various applications.
\let\thefootnote\relax\footnotetext{$^{\dagger}$ denotes equal contributions and authors listed alphabetically.}

However, there is a significant gap between the research on these models using powerful development hardware and their performance and applicability on resource-constrained hardware available in practical platforms. 
There are several challenges to their practical deployment. 
For instance, real-world platforms like robots, autonomous vehicles, and mobile devices are constrained by available onboard power, memory, and computation while working in real-world scenarios.
In this paper, we shall work towards addressing three such challenges and propose techniques that can enable the deployment of these models to practical scenarios.

First challenge comes about primarily due to the large size of these models, making them tedious to use in practice because of the resource constraints on the real world onboard platforms. 
This brings in a need for smaller and lightweight models that perform as well as the larger and deeper models or at-least take a minimal hit in performance. 
This is referred to as model compression problem and is an active research area where several successful techniques have been proposed. 
One of which is the method of Knowledge Distillation (KD), which involves training a smaller model with supervision from the larger model or an ensemble of models while obtaining good performance.

The second challenge is that during the development phase, the neural networks are trained on a known set of classes and tested on the same. 
Whereas, in practical scenarios, a model would encounter samples from unknown classes as well, which were not a part of the training classes, also referred to as the closed set samples. 
For instance, a robot or an autonomous vehicle cannot be trained for every possible object class that it may come across but still needs to work in real-world conditions with several such open set samples.
Moreover, due to the design of standard neural network architectures used, these models would still classify the unknown class samples to one of the known classes with high confidence. 
This is an undesirable outcome in practice as we would ideally want the model to separate the known classes from unknown classes while also correctly classifying the known ones.
This is referred to as the Open Set Recognition (OSR) problem, which is comparatively a more nascent research area.
A model's OSR capability makes it better suited for practical use by making it robust to unknown class samples encountered in practice.

The third challenge is that most computer vision research efforts focus on working with standard visual datasets involving images and videos. 
Deep learning models have been widely researched and have significantly improved image recognition and other vision tasks. 
However, the real world is three dimensional, and a common shortcoming of these models is that they process 2D projections of the 3D objects and ignore the actual 3D nature of the objects and scene.
Exploring this 3D information by extending these models to 3D would improve the performance of various practical applications like autonomous driving and augmented reality while also improving original 2D image recognition tasks. 
However, in spite of the growing interest, this area has been relatively under-explored in research.

Moreover, despite their practical applicability, KD and OSR have been developed independently in the literature. 
In this work, we aim to bridge these two techniques such that a student model acquires knowledge of the teacher model while becoming robust to open set samples by joint OSR training, increasing its practical utility. 
This paper shows how the KD process by itself can transfer open set capabilities to a student network along with the previously known dark knowledge transfer.
Furthermore, we present an open set training methodology to further enhance the distilled student model's OSR performance.
Existing works in the KD and OSR literature employ only text and regular media, this work also aims to show how we can apply these to 3D objects. 
Our primary contributions in this work are:
\begin{itemize}
    \item We propose a method to perform knowledge distillation from a large teacher model to a smaller student model while simultaneously training the student network for open set recognition to improve its robustness.
    \item We propose a novel loss objective and a joint training methodology for KD and OSR.
    \item We demonstrate the effectiveness of the proposed method using detailed experiments on 3D object data.
    \item We study how this leads to a tradeoff in distillation and open set performance of the learned student network.
\end{itemize}

\section{Literature Review}
There are many popular representations for 3D data, such as voxels, meshes, and point clouds, and each of them has different approaches for learning and recognition. 
In this work, we have focused on point cloud representation, and here we discuss some of the key advances in deep learning techniques for this task.
One of the pioneering works is by \cite{qi2017pointnet}, where the authors make direct use of point clouds as an input to a deep network.
Subsequently, the same authors proposed another model called PointNet++ \cite{qi2017pointnet++}, which applies PointNet recursively to aggregate local information and pass it to the next step for improved performance.
\cite{li2018pointcnn} presented an architecture where they learn a $\chi$-transformation, to generate features from input point clouds on which convolutions can be directly applied.
More recently, \cite{liu2019relation} proposed relation shape CNN, where the network learns to exploit the geometric topological constraints among the 3D points for improved performance.
For our work, we use the PointNet architecture by \cite{qi2017pointnet} as the base model and use it to demonstrate KD and OSR on point cloud data for the first time in the literature to the best of our knowledge.

Distilling knowledge from a large trained machine learning model to a smaller model was first introduced by \cite{bucilua2006model}.
The authors proposed to minimize the square difference between the logits obtained from the larger and smaller models and demonstrated their method both for distilling knowledge from a large model and an ensemble of models.
More recently, \cite{hinton2015distilling} developed a more general approach within the scope of neural networks and for handwritten digit recognition.
They scale the network logits by a temperature parameter to soft thresholds that are matched for distillation, and show the work in \cite{bucilua2006model} is a special case of their method. 
In generative adversarial networks based approaches, the student mimics the teacher as a generator mimics the real data distribution \cite{wang2018kdgan}, \cite{xu2017training}, \cite{zhang111adversarial}, \cite{liu2019exploiting}.
Recently, \cite{tian2019contrastive} developed the contrastive representation distillation, which uses contrastive learning to match the representations of the teacher and student networks.
This is currently the state of the art method for KD and, we build upon this work in our paper.
OSR is comparatively a recent research area, and several works in this field adopted a threshold-based classification scheme for classifying unknown classes.
\cite{scheirer2014probability} formulated a compact abating probability model for this problem, which uses two Support Vector Machines (SVMs). 
Here, the first SVM identifies if the input sample is from unknown class, depending on if the posterior estimate of an input sample falls below the threshold. 
Otherwise, the sample is passed to the second SVM for further classification into one of the known classes.
\cite{zhang2020hybrid} recently incorporated this idea of threshold-based classification into the paradigm of deep networks for OSR.
Further, \cite{sun2020conditional} and \cite{yoshihashi2019classification} also use a softmax model augmented with a threshold probability as a baseline model for their experiments and we too build on similar directions.
However, as opposed to the previous methods, we propose a novel technique to perform both KD and OSR in a joint manner.

\section{Proposed Method}
In this work, we distill a larger teacher network's knowledge to a smaller student network while making it more robust by jointly performing open set training for a 3D object recognition.

\subsection{Dataset Preparation}
In this paper, we work with 3D point cloud data, unlike existing KD and OSR literature, which is focused on using a regularly spaced grid-like datasets like audio, images, and videos. 
A point cloud is a set of points in Cartesian 3D space \{$P_i | i = 1,2,...,n$\} where each point $P_i$ is a vector having $(x,y,z)$ coordinates. 
There are other feature channels as well, such as color, material, and texture; however, we do not consider them in this work. 
In order to work with this 3D point cloud data, we follow the data preparation approach used by the initial work in the PointNet paper \cite{qi2017pointnet}. 
Here, we first sample a set of $N_0$ points from the total number of points of a point cloud and reshape them to an $N_0$ length vector, $x$. 
This vector $x \in R^{N_0 \times 3}$ then represents our point cloud data sample and can now be handled by the proposed deep network architectures.

\subsection{Network Architecture}
\label{sec:netArch}
\begin{figure*}[t]
    \centering
    \includegraphics[scale=0.4]{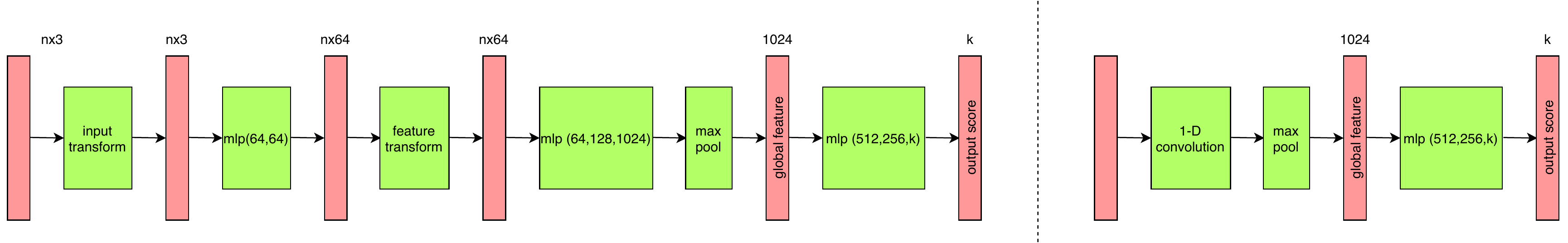}
    \caption{The architecture of the Teacher (left) and the student (right) networks used in our experiments, which are inspired by the original PointNet architecture \cite{qi2017pointnet}.
    The architecture of the Student network constructed by removing the transform layers along with two multi layer perceptron blocks from the Teacher model.
    Teacher parameter count = 3463763.
    Student parameter count = 666378 (Only 19\% parameters as the Teacher model).}
    \label{fig:netArchs}
\end{figure*}

We use the PointNet network architecture proposed by \cite{qi2017pointnet} as our base model i.e., our teacher model. 
This architecture's main components are the input transform and feature transform, symmetry function for unordered input, and the final classification by the softmax layer.
For point cloud classification, the learned representations should be invariant to transformations like rotation and translation of all the point clouds. 
To impose this condition, PointNet uses an input transform layer to predict a $3\times3$ transformation and directly applies it on the input point set, which is a $n\times3$ matrix. 
They have also imposed a similar condition on the alignment of feature space at an intermediate stage performed by a sub-module named feature transform. 
However, since the dimensionality of the feature space is much higher than that of the coordinate space, the authors have added a regularization term to the loss function.
A schematic of the PointNet architecture is presented in Figure \ref{fig:netArchs} along with the student network used in our experiments.

The other key component of the PointNet architecture is the symmetry function for unordered input. 
Unlike image datasets, the points in a point cloud are unordered and the model needs to be unaffected by input permutations. 
In order to make the model invariant to input permutations, authors use a symmetric function on the features of a point cloud.
A symmetric function takes $n$ input vectors and outputs a vector that is invariant to input order. For example, in case of 2 input vectors, + (vector addition) and $\cdot$ (dot product) are symmetric functions.
The symmetric function used in their work is a max pool function.
After the feature generation, they are passed to a multi-layer perceptron and a softmax layer that generates probabilities of an input sample lying in different classes.
The student model used for our experiments is constructed by removing the input transform and feature transform from the teacher model architecture, along with some more layers from the PointNet network.
The number of parameters in the student network is about 19\% of the teacher network leading a $\sim5\times$ compression.

\subsection{Distillation Methodology}
\label{sec:KD}
For KD, we first train a large teacher network and smaller student network, both of whose architectures are described in the previous section \ref{sec:netArch}. 
The performance of this teacher-student pair trained from scratch would act as our baseline for further experiments and performance evaluation.
We propose to perform KD by building on the previous works by \cite{hinton2015distilling} and contrastive representation distillation by \cite{tian2019contrastive}. 
Additionally, as we have the original labels available, we adopt the approach by \cite{hinton2015distilling} where the authors propose to train the student network using the logits obtained from the teacher network, in addition to the original labels.

Let us formalize this by introducing some notation for clarity.
Let $x \sim p_{data}(x)$ denote and input sample coming from a data distribution.
Furthermore, let $f^S$ and $f^T$ be the functions denoting the student and teacher networks until their penultimate layer.
The output of these networks are the penultimate features, i.e. $S$ and $T$ respectively and can be written as,
\begin{equation}
 S = f^S(x)
\end{equation}
\begin{equation}
 T = f^T(x)
\end{equation}
Following which we have the final layers of the Teacher and Student networks which are denoted by the functions $g^S$ and $g^T$, respectively.
These layers are essentially a fully connected layer and the final teacher and student network logits are represented as $z^S$ and $z^T$ and can be defined as follows,
\begin{equation}
 z^S = g^S(x)
\end{equation}
\begin{equation}
 z^T = g^T(x)
\end{equation}
Finally, the probabilities of student and teacher networks are given by $\sigma(z^S)$ and $\sigma(z^T)$, where $\sigma$ denotes the softmax function.
To summarize the notations, the complete student network is represented in terms of the composite function $\sigma^S(g^S(f^S(x)))$.
Similarly, the complete teacher network can be written as $\sigma^T(g^T(f^T(x)))$.

We first train our baseline distilled student model using the KD technique, as proposed in \cite{hinton2015distilling}.
This student is trained using the soft logits from the teacher network while minimizing the following loss objective,
\begin{equation}
 \mathcal{L_{KD}}(x,W) = L_{CE}(\sigma(z^S;\tau_{KD}); \sigma(z^T;\tau_{KD}))
\end{equation}
where, $W$ are the parameters of the student model, $L_{CE}$ is the cross entropy loss and $\tau_{KD}$ is the temperature parameter used for calculating the soft logits.
The soft logits are obtained using the modified softmax function with a temperature parameter as shown in Equation \ref{eq:softLogits}, initially proposed by \cite{hinton2015distilling}.
\begin{equation}
 \sigma(z,\tau_{KD}) = \frac{exp(z_i/\tau_{KD})}{\sum_{j}^{} exp(z_j/\tau_{KD})}
 \label{eq:softLogits}
\end{equation}
where the summation is over all the classes.

Next, we use the contrastive representation distillation method proposed by \cite{tian2019contrastive}, which proposed to maximize the lower bound on the mutual information between the teacher and student representations. 
To achieve this, they learn a representation that brings the positive pairs closer and negative ones farther in the feature space, using the loss function,
\begin{multline}
 \mathcal{L}_{CRD}(h) = \mathbb{E}_{q(T,S|C=1)}[log \textit{ h}(T,S)] \\ + N\mathbb{E}_{q(T,S|C=0)}[1-log \textit{ h}(T,S)]
\end{multline}
where, $q$ is a distribution with latent variable $C$ which represents whether ($f^T(x_i)$,$f^S(x_i)$) belong to the joint $(C=1)$ or the product of marginals $(C=0)$.
\begin{align} 
 q(T,S|C=1) &= p(T,S) \\ 
 q(T,S|C=0) &= p(T)p(S)
\end{align}
And $h(T,S)$ is a mapping of $q(C=1 | T,S)$ as given below,
\begin{equation}
 \textit{h}(T,S) = \frac{exp(G^T(T)^{'}G^S(S)/\tau_{CRD})}
 {exp(G^T(T)^{'}G^S(S)/\tau_{CRD}) + \frac{N}{M}}
\end{equation}
where, $M$ is cardinality of the dataset, $\tau_{CRD}$ is a temperature parameter used for CRD loss term and $G^T$ and $G^S$ linearly transform $S$ and $T$ into same dimension, taking care of the differences in dimensionality of $S$ and $T$, if any, and further normalize them by $\mathcal{L}$2 norm before the inner product.
Building on these, we use a weighted combination of the above three losses in our proposed distillation procedure for the student model.

\begin{figure*}[t]
    \centering
    \includegraphics[scale=0.58]{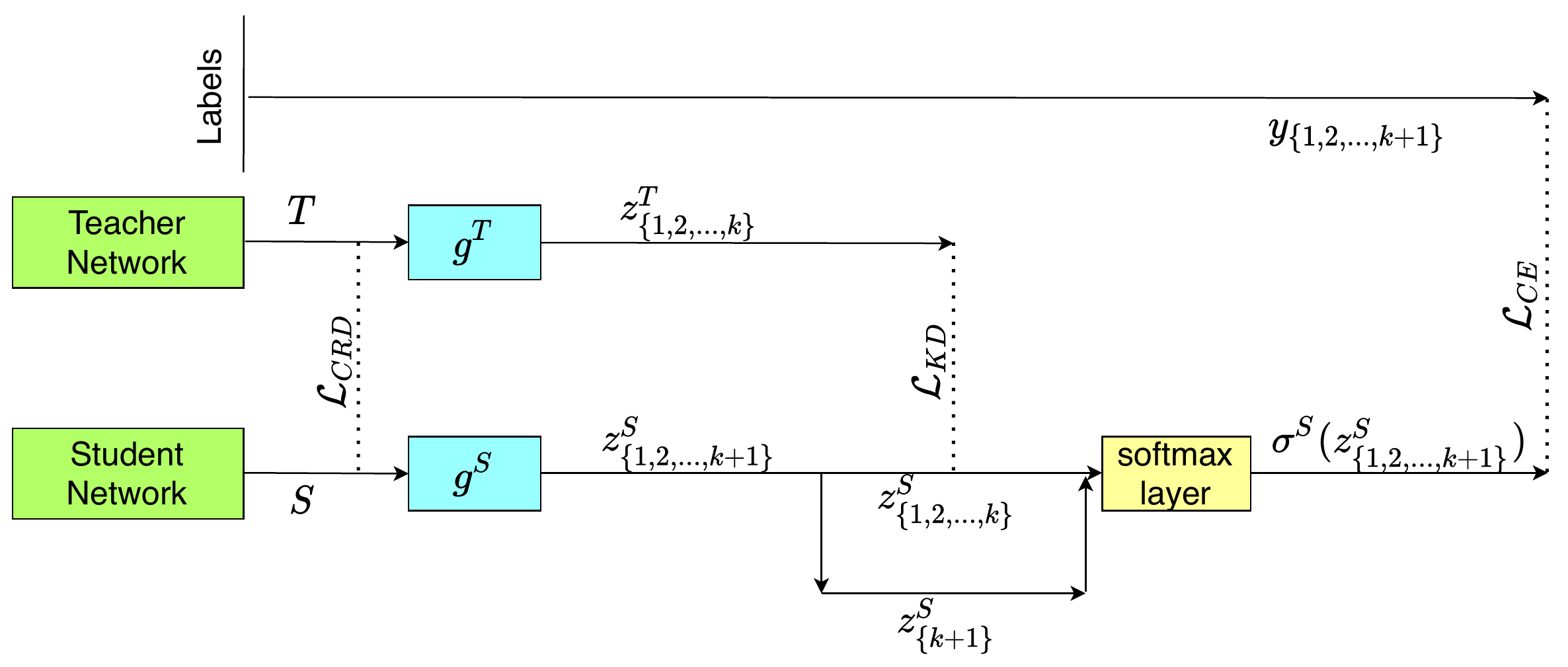}
    \caption{A schematic of the proposed methodology for joint knowledge distillation and open set recognition.
    The outputs from teacher and student models are their respective representations, T and S, which are used for calculating CRD loss term.
    Next, we $z^S$ and $z^T$ represent student and teacher logits, which are used for calculating the KD loss term.
    Finally, we calculate class probabilities $\sigma^S(z^S)$, which are used for computing of cross-entropy loss term.
    }
    \label{fig:jointTraining}
\end{figure*}

\subsection{Open Set Recognition Methodology}
\label{sec:OSR}
OSR deals with the problem of training a classifier using $k$-known classes and testing it on a dataset containing samples from these known classes as well as samples not belonging to any of the known classes. 
The goal here is that the classifier should reject the unknown samples while classifying the known samples into their respective $k$ classes. 
Following the performance evaluation in previous works, our baseline OSR model consists of weights of a softmax classifier trained only on known classes with a threshold probability. 
As per this, during inference, if the maximum probability of an input to lie in any known class is less than this threshold probability, the model classifies it as an unknown sample. 
Otherwise, the sample is classified into one of the known classes. 
Mathematically stated,
\begin{equation}
 pred(x)=\begin{cases}
 (k+1), & \text{if $p_{max} < p_{threshold} $}\\
 p_{max}, & \text{otherwise}
 \end{cases}
\end{equation}
where $p_{max} = arg max_{j \in \{1,...,k\}} p(y_j|x)$.

While analyzing the OSR performance of a model, we must consider both the rejection accuracy, which is the accuracy of the model on the $(k+1)$-th class, i.e., the unknown classes bucket, and the classification accuracy, which is the accuracy of the model on known classes. 
A more detailed analysis of selecting this threshold probability has been explained in the results and analysis section. 
As part of the proposed method, to reject the unknown samples, we need to train the model on samples from all the known classes and some samples not belonging to any of these known classes. 
We achieve this by generating new samples from the existing known class data, such that they do not belong to any of the known classes.
We refer to these as the pseudo open set samples used for OSR training, and they contain point clouds formed by mixing a pair, triplet, and quadruplet of point clouds belonging to different known classes. 
The algorithm to do this is described in Algorithm \ref{alg:sampleGeneration}.

\begin{algorithm}[t]
 \SetAlgoLined
 \For{\textnormal{n in \{2, 3, 4\}}}{
 \For{\textnormal{i in \{0, $N_n$\}, where $N_n$ is the number of generated samples}}
 {- Randomly select $n$ point clouds from different classes\;
 - Stack the coordinates of $n$ selected samples\;
 - Shuffle the triplets of $(x, y, z)$ coordinates in random order\;
 - Equally separate the shuffled set of point sets \;
 - Then, form $n$ new point clouds which won't belong to any of the classes\;}
 }
\caption{Proposed pseudo-open set sample generation}
\label{alg:sampleGeneration}
\end{algorithm}

We have done a more detailed analysis to show that these newly formed point clouds do not belong to any known classes using TSNE plots of the feature space. 
Using the proposed pseudo open set sample generation strategy, we train our models for OSR using a classification loss objective.

\subsection{Choice of Threshold parameter}
As explained above, we augment our softmax classifier with a threshold probability and use it as our baseline for OSR evaluation. 
This threshold is chosen as 0.5 as the model's OSR performance should not significantly affect its closed set performance.
As we can see in Table \ref{tab:OSR_results}, the baseline model obtained by augmenting our student trained from scratch, the classification accuracy drops only by 2.04 \% by adding a threshold. 
As the threshold increases, the model's closes set accuracy suffers a lot even though the overall accuracy increases. 
We adopt this approach from previous works \cite{sun2020conditional} and \cite{yoshihashi2019classification} that also use 0.5 threshold baseline.

\subsection{Joint Open Set Distillation Methodology}
We incorporate the ideas of KD and OSR into one model for our final setup and propose a joint training methodology and a suitable loss objective. 
For this, we take pre-trained teacher network trained only on $k$ known classes and append the known class data using the proposed pseudo open set sample generation strategy introduced in the previous section.
The final layer of the student model, which is much smaller in size compared to the teacher network, is therefore modified to be a $(k+1)$-way classifier to suit this problem.
This allows us to use the proposed KD and OSR training techniques proposed in the previous two sections \ref{sec:KD} and \ref{sec:OSR}.
In addition, we use the proposed pseudo open set sample generation process for OSR training.

The complete architecture and training process of this proposed joint KD and OSR methodology are presented in Figure \ref{fig:jointTraining} for clarity.
To achieve both distillation and OSR, our proposed loss objective consists of three terms, a cross-entropy loss term along with KD loss introduced by in \cite{hinton2015distilling} and CRD loss term as introduced in \cite{tian2019contrastive}. 
We use all the 11 logits generated from the $k+1$-classifier of the student network to compute the cross-entropy loss. 
While calculating the other distillation loss terms, we only use 10 logits corresponding to the $k$ known classes. 
This way, we are training a smaller and more robust network for OSR in a single step without significantly impacting its distilled closed set performance.

\section{Experiments and Results}
\subsection{Dataset Description}
\label{sec:dataset}
For our experiments, we use the ModelNet10 and ModelNet40, which are the standard 3D object recognition datasets in the literature.
ModelNet40 \cite{wu20153d} comprises of a total of $12311$ CAD models from $40$ man-made object categories, split into $9843$ samples for training and $2468$ samples for testing.
The ModelNet10 dataset, on the other hand, is a 10 class subset of ModelNet40.
The train to test split for ModeNet10 is $3991$ to $904$, divided into $10$ classes.
Since the number of points varies from one point cloud to another, to keep the input size uniform, we sample $1024$ points from each point cloud, making the input of uniform shape $1024 \times 3$.
We essentially use $N_0$ = 1024, and the rest of the explanation is the same as in section 3.1. In order to use these datasets for both KD and OSR analysis, we split them into closed and open set classes to obtain two datasets to work with.
From the ModelNet40 dataset, we select 10 classes of ModelNet10 as the known classes, representing the closed set classes.
The samples belonging to the remaining 30 classes are used as unknown class or open set samples.
We use this dataset for all our experiments and evaluation.

\begin{table}[t]
    \label{tab:KD_results}
    \centering
    \caption{Performance of various student models using different Knowledge Distillation procedures as described in Section \ref{sec:KD}.}
    \begin{tabular}{|c|c|c|}
    \hline
    Models & Loss term & Accuracy \\ \hline
    Teacher & CE & 93.67 \\ \hline
    Original Student & CE & 87.17 \\ \hline
    & CRD+CE & 88.03 \\
    Distilled Student & KD & 88.47 \\
    & CE+KD & 87.6 \\
    & KD+CRD+CE & \textbf{88.69}\\
    \hline
    \end{tabular}
\end{table} 

   \begin{table*}[t]
    \centering
    \caption{Open Set Recognition performance of the proposed distilled open set model along with only distilled and only OSR trained models along with teacher-student pair baselines for comparison.}
    \label{tab:OSR_results}
    \begin{tabular}{|c|c|c|c|c|}
        \hline
        Model & F-measure & Total Accuracy & Acc closed set & Acc open set \\ \hline
        Teacher & 37.89 & 48.36 & 93.39 & 7.07 \\ \hline
        Scratch Student & 33.47 & 43.62 & 85.13 & 5.55 \\
        Proposed Distilled Student & 38.37 & 46.78 & \textbf{87.44} & 9.49 \\
        Proposed Open Set Student & 46.04 & 50.79 & 85.68 & 18.79 \\
        \hline 
        Distilled Open Set Student & \textbf{43.97} & \textbf{49.89} & 86.78 & \textbf{16.06} \\
        \hline 
    \end{tabular}
\end{table*}

\subsection{Experimental Setup}
We have done all our experiments in python using PyTorch deep learning framework.
Our models take around 30-60 minutes to train and converge to a solution on an Nvidia GeForce RTX 2060 GPU.
We use the PointNet architecture as our base model for experiments, which acts as the large teacher network.

\subsection{Knowledge Distillation Performance}
We first independently train a large Teacher network and a smaller student network for classification, using only the closed set samples of the data.
The performance of these the large teacher and the smaller student models provide us with the baseline performances to evaluate the proposed KD methodology.
We perform KD using supervision from this teacher model onto a student model using the distillation methodologies described in Section \ref{sec:KD}.
For our final closed set distilled student model we use the proposed KD methodology and train with a weighted combination of the three loss terms, leading to the final distillation loss objective is as follows:
\begin{equation}
 \mathcal{L} = \alpha\mathcal{L}_{KD} + \beta\mathcal{L}_{CRD} + \gamma\mathcal{L}_{CE}
\end{equation}
where the hyperparameters $\alpha$, $\beta$ and $\gamma$ are obtained using grid search.
We use the overall accuracy as our performance metric, and the results of these experiments are presented in Table \ref{tab:KD_results}.

We can observe from the table that the distilled student performs best when all three loss terms have been used.
We also found that the original KD procedure is more effective than the plain CRD method in our setup.
However, using both KD and CRD helps improve student performance compared to using just either one.
Moreover, the proposed method provides the best distilled student performance, and overall, there is a significant improvement from our baseline student model.
This demonstrates how using distillation helps training a better performing, smaller student network than trained from scratch.

\subsection{Open Set Recognition Performance}
For the proposed OSR training, we require representative unknown training samples, which we refer to as pseudo open set samples, that are not part of the closed set classes.
For this, we first use the proposed method to generate such pseudo open set samples using only the closed set samples, details of which are described in Section \ref{sec:OSR} and Algorithm \ref{alg:sampleGeneration}.
With the new train and test data prepared from ModelNet40, the teacher network used is the same as that of the non-open set part, but the loss objective has been modified, as explained in the previous section \ref{sec:OSR}. 
Finally, we test the open set capabilities of this model using the open set class samples of our dataset.
To train our final model, which has fewer parameters and improved open set capabilities, we distill the knowledge from the teacher to the student network.
The student model trained from scratch and the distilled student model from the KD step above are our two baselines that we compare the open set performance with.

The results of this experiment are reported in Table \ref{tab:OSR_results}.
We use F-measure to evaluate the open set performance of the proposed model and baselines.
From the table, we can observe that the student model trained from scratch is much lower in all the four performance metrics, which is expected due to its smaller size and the standard classification training process employed.
Along with this, we can observe that the teacher network has a higher closed set and open set classification performances due to its larger parameter count.
Additionally, we observe that the proposed open set student model achieves a better performance than the student from scratch and the teacher model itself.
Interestingly, there is also a slight improvement in the open set capabilities of the distilled student model, which we attribute to the supervision recieved from the teacher network during the proposed distillation training.
This finding implies that the along with the transfer of dark knowledge from teacher to student as proposed by \cite{hinton2015distilling}, distillation process also transfers open set capability to the distilled student.

\subsection{Joint Distillation and Open Set Recognition Performance}
Finally, we evaluate joint distillation and open set training of the student network using the proposed training approach.
The results of these experiments are also presented in Table \ref{tab:OSR_results}.
We compare the performance of the proposed joint training procedure to the proposed distillation and OSR training performances along with normal teacher and student training.

We can observe that, when we incorporate the proposed joint KD and OSR training process, we observe a significant jump in the student model's open set capabilities as reflected in the four performance metrics in Table \ref{tab:OSR_results}.
For instance, we see over a 5\% increase in F-measure and over 3\% in overall accuracy compared to the student obtained using proposed distillation.
Similarly, the accuracy on the open set classes has also gone up by about 7\%.
However, there is a slight drop in the accuracy on closed set classes.
We would like to highlight that, in our experiments, we observe a trade-off that comes into play between the open set performance of a distilled model and its closed set performance, increasing one leads to a drop in the other.

\subsection{Ablation Study}
\subsubsection{KD Temperature ($\tau_{KD}$)}
\begin{figure}[t]
    \centering
    \begin{tabular}{cc}
        \includegraphics[scale=0.3]{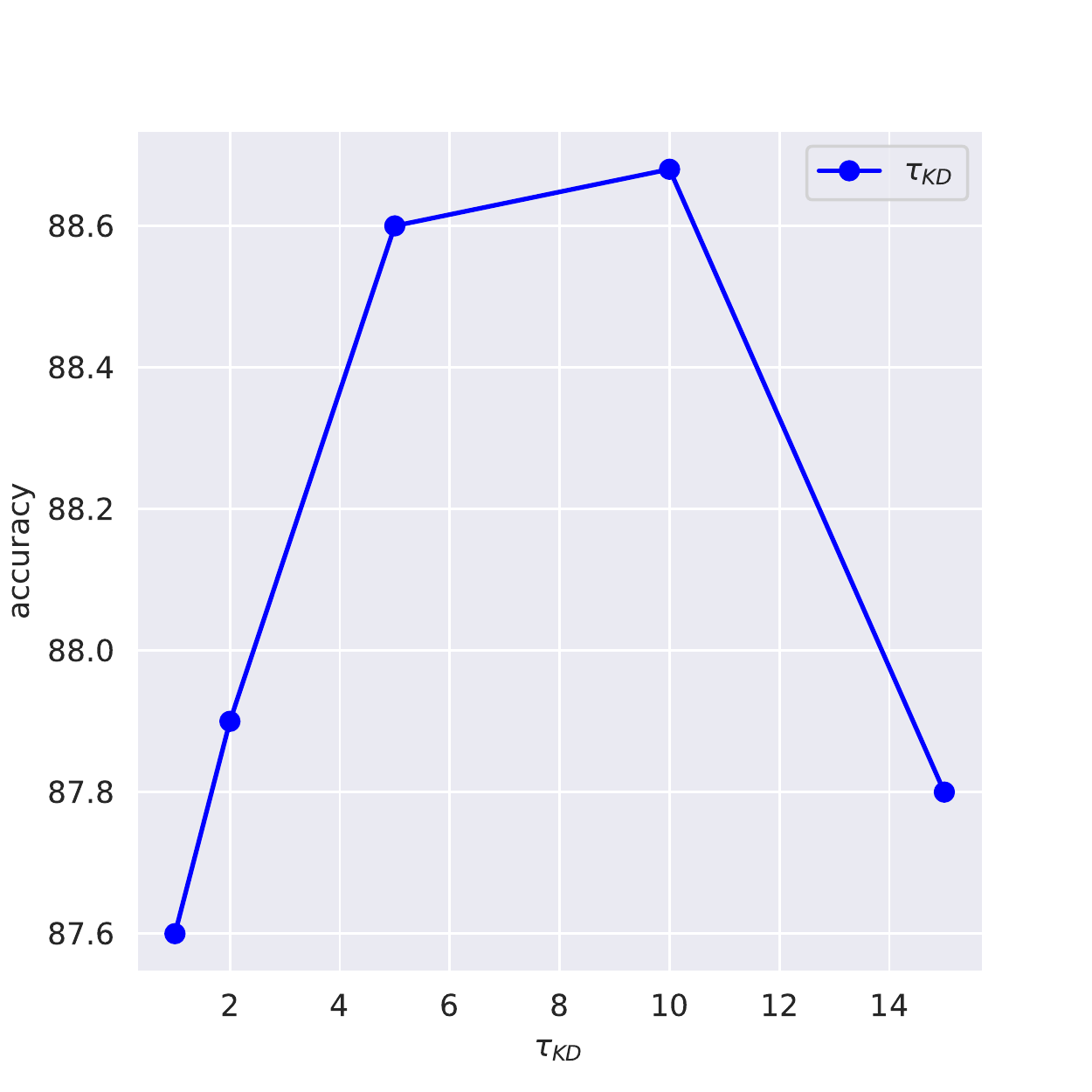} & 
        \includegraphics[scale=0.3]{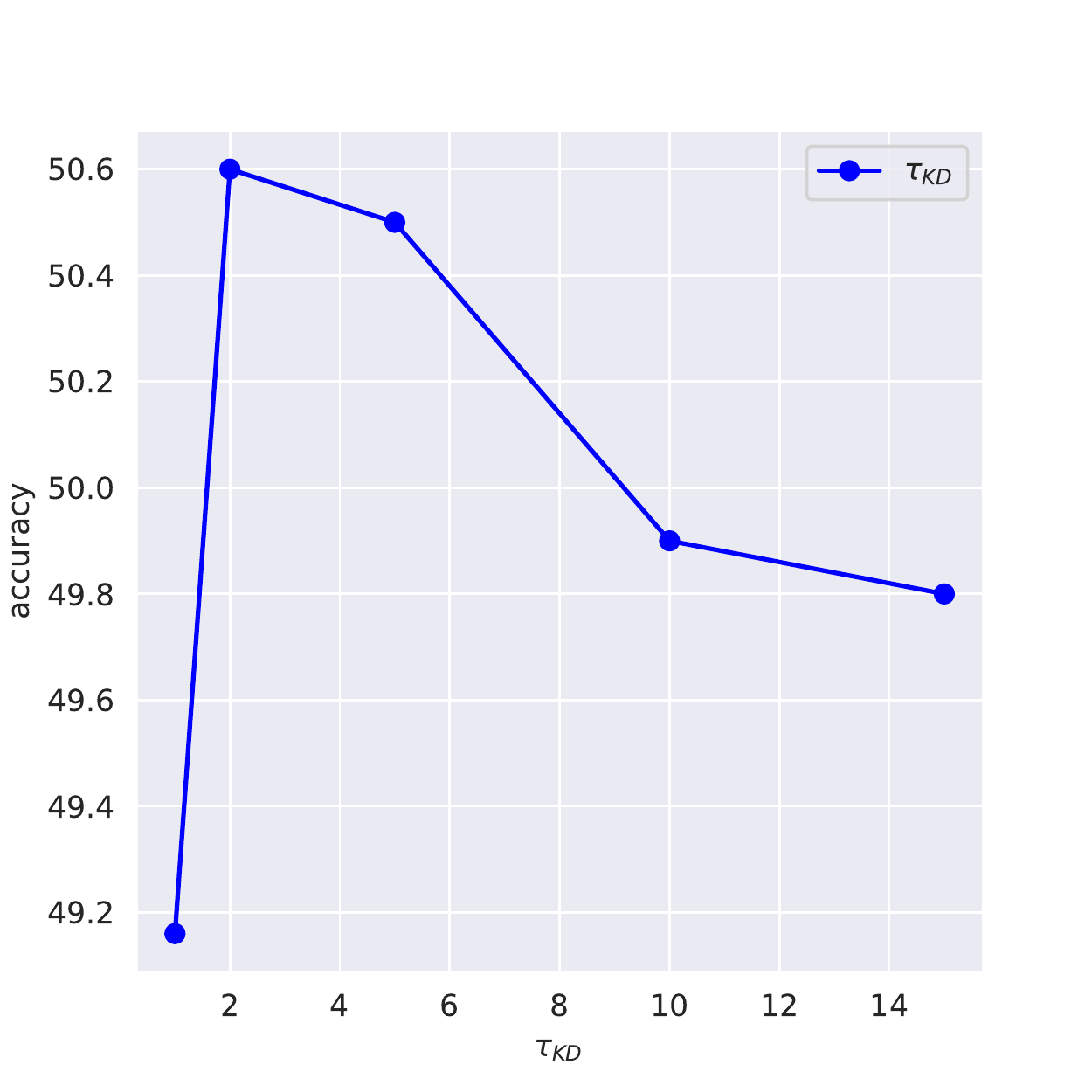} \\
        (a) & (b) \\
    \end{tabular}
    \caption{Plots for (a) closed and (b) open set performance variation by changing the Temperature parameter used for Knowledge Distillation.}
    \label{fig:T_KD_plots}
\end{figure}

Here, we study the effect of varying the temperature parameter for KD training.
The results of this are presented in Figure \ref{fig:T_KD_plots}.
We can observe from these plots that the distilled student network's performance peaks at a specific temperature and then falls off at lower and higher temperatures.
Hence, we choose the best temperature parameter for KD as 10 to obtain the best distillation performance.
We observe that a analogous trend is evident in the OSR performance, and we choose $T_{KD} = 10$ for the final distilled open set student model for a good trade-off.

\subsubsection{CRD Temperature ($\tau_{CRD}$)}
\begin{figure}[t]
    \centering
    \begin{tabular}{cc}
        \includegraphics[scale=0.3]{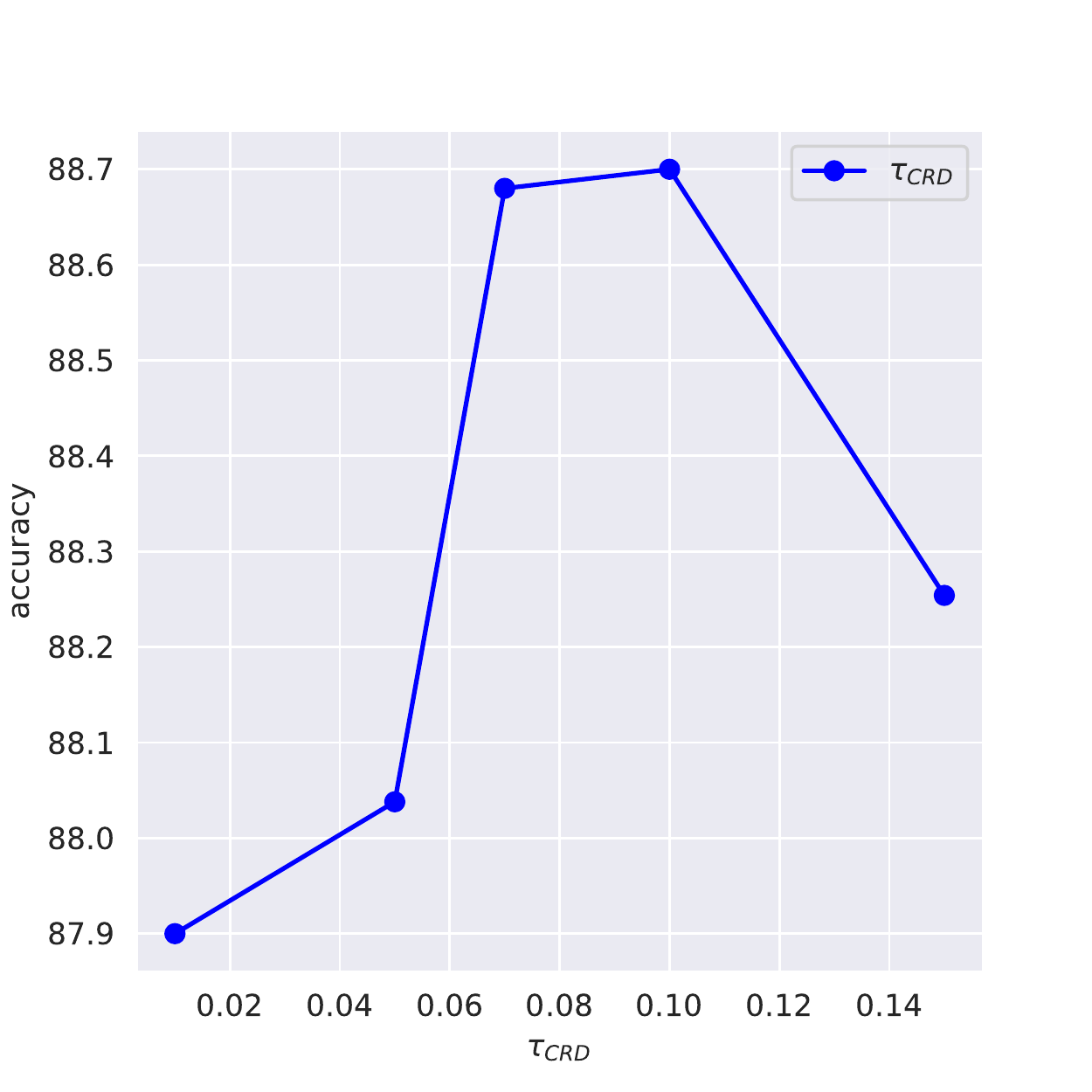} & 
        \includegraphics[scale=0.3]{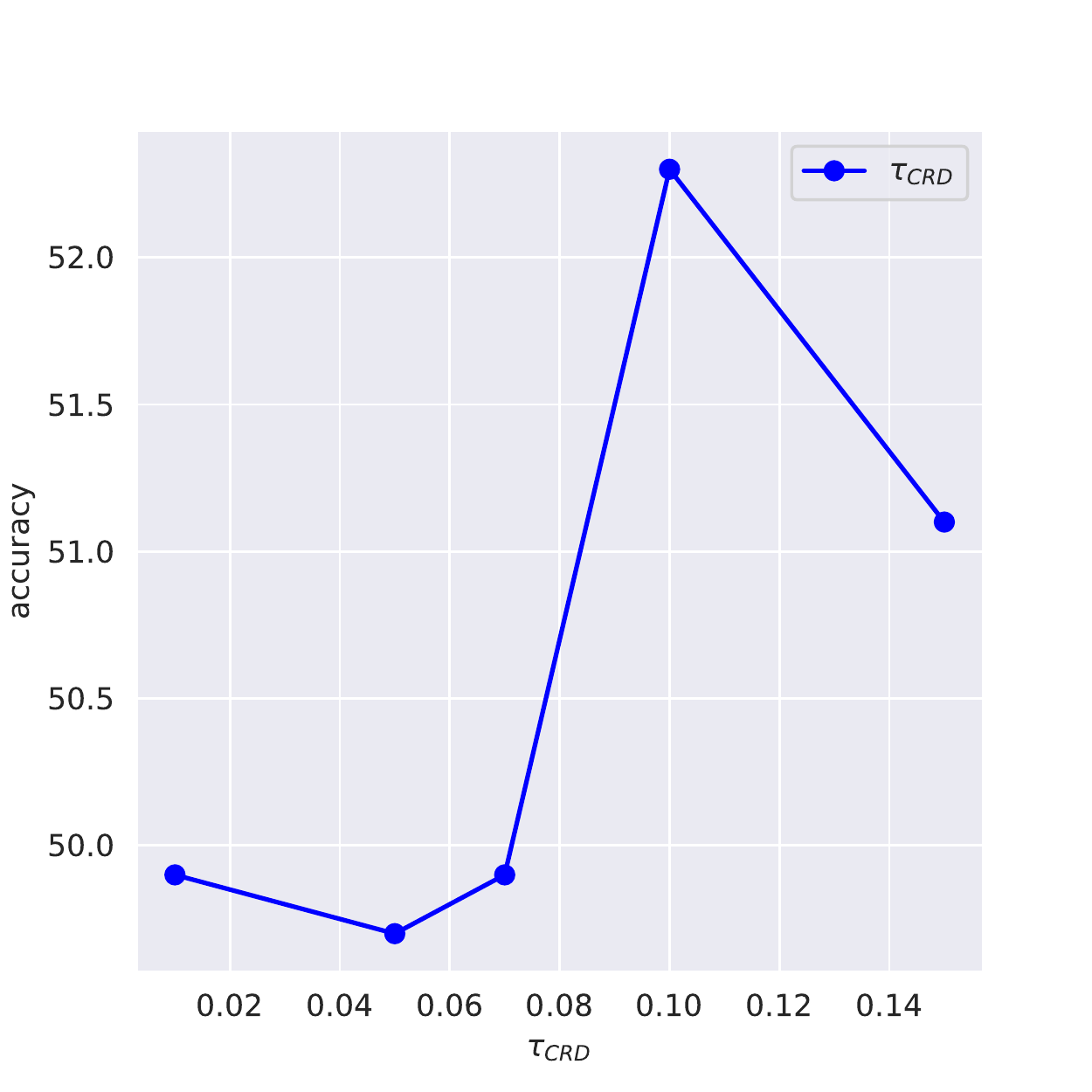} \\
        (a) & (b) \\
    \end{tabular}
    \caption{Performance variation for (a) closed and (b) Open set by changing the Temperature parameter used for Contrastive Distillation procedure.}
    \label{fig:T_CRD_plots}
\end{figure}

We also study the performance variation of the student model by varying the temperature parameter used for the Contrastive distillation procedure.
We observe a similar trend in performance as in the previous experiment for the variation of the KD temperature parameter.
The results of this are presented in Figure \ref{fig:T_CRD_plots}.
We can observe that the performance is low at high and low temperatures and peaks at a particular temperature value.
We choose these peak values as our final parameters i.e., 0.10 for the final distilled model and 0.10 open set distilled model.

\begin{figure}[t]
    \centering
    \begin{tabular}{cc}
        \includegraphics[scale=0.08]{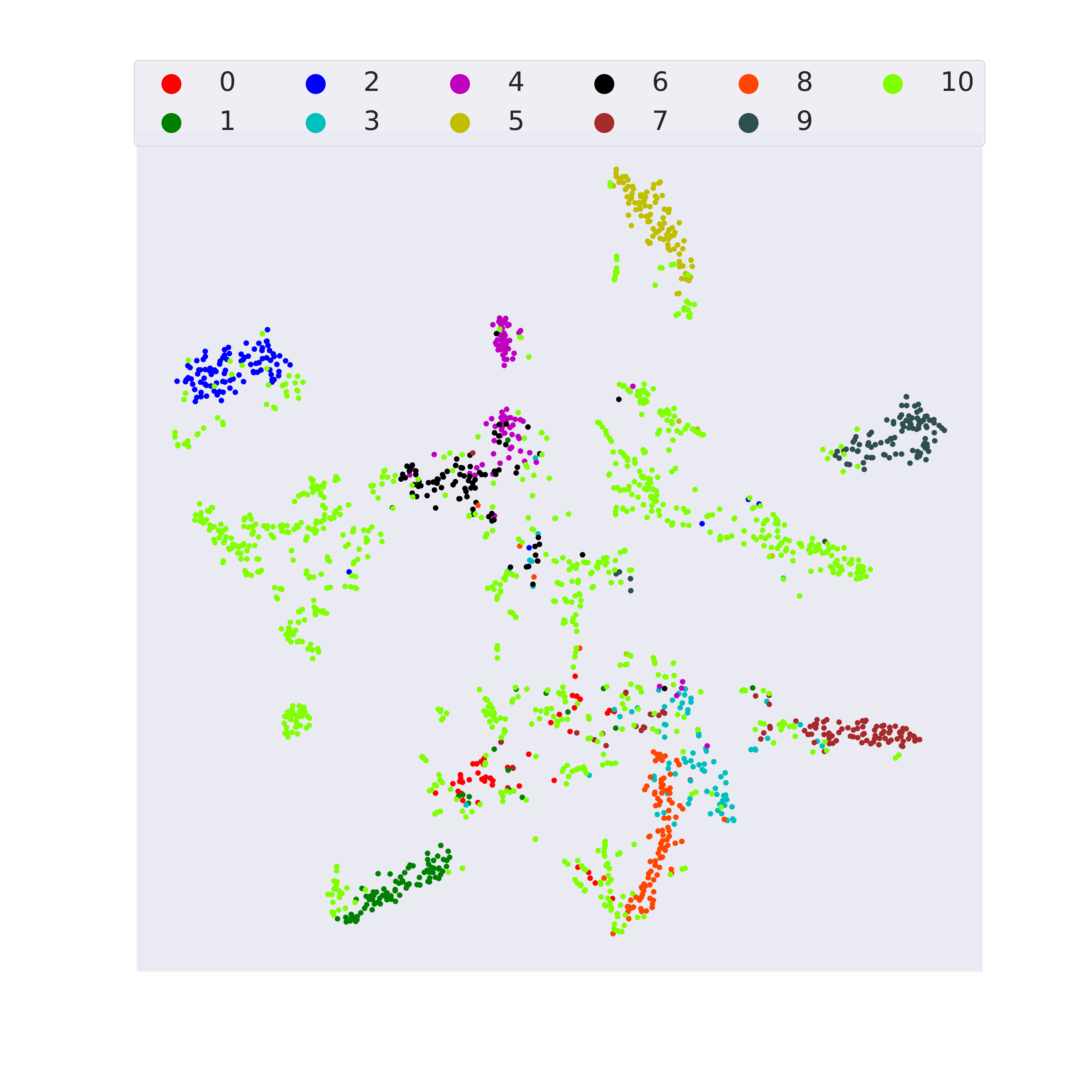} & 
        \includegraphics[scale=0.08]{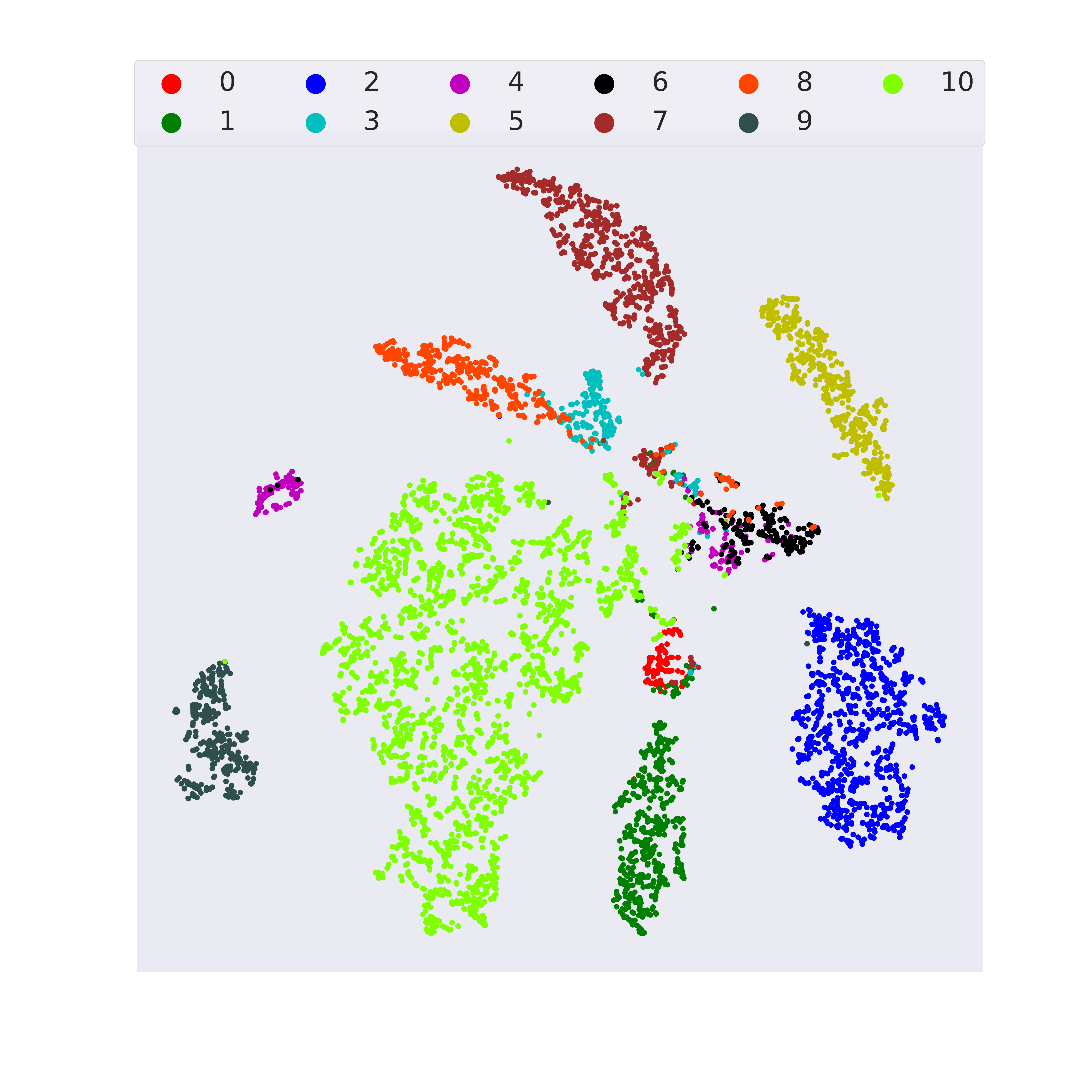} \\
        (a) & (b) \\
    \end{tabular}
    \caption{(a) TSNE plots for the closed set class sameples along with the open set samples used for testing.
    (b) TSNE plots for the closed set samples along with the generated pseudo open set samples used for training.}
    \label{fig:TSNE_plots}
\end{figure}

\subsubsection{Latent Domain Visualization}
We present TSNE plots of the open and closed set samples in Figure \ref{fig:TSNE_plots} to understand their distribution.
In the first plot, we can observe how the closed set samples are clustered and how the open set samples are distributed across this feature space.
The second plot shows how the open set sample we generate using the proposed technique are distributed.
The class labels in the figure legend from 0-9 represent closed set samples, and the marker 10 represents the open set samples.
This demonstrates the effectiveness of our proposed pseudo-open set sample generation and how the generated samples belong to a cluster different from the known classes, benefiting the OSR training.

\section{Conclusion}
This paper introduces a novel joint training methodology for knowledge distillation and open set recognition for three-dimensional point cloud recognition.
We propose a sample generation process to generate pseudo open set samples for OSR training of the student network.
The proposed method's effectiveness is presented using a detailed evaluation on the standard ModelNet 3D point cloud data.
We also present two interesting findings from our experiments in this work.
First, knowledge distillation also facilitates OSR capability transfer from a teacher network to student network apart from the previously known dark knowledge for closed set recognition.
Secondly, while performing joint distillation and open set training, we observe a trade-off between the closed set and open set performances of the final student model.

\bibliographystyle{IEEEbib}
\bibliography{refs}

\begin{thebibliography}{10}

\bibitem{qi2017pointnet}
Charles~R Qi, Hao Su, Kaichun Mo, and Leonidas~J Guibas,
\newblock ``Pointnet: Deep learning on point sets for 3d classification and
  segmentation,''
\newblock in {\em IEEE conference on computer vision and pattern recognition},
  2017, pp. 652--660.

\bibitem{qi2017pointnet++}
Charles~Ruizhongtai Qi, Li~Yi, Hao Su, and Leonidas~J Guibas,
\newblock ``Pointnet++: Deep hierarchical feature learning on point sets in a
  metric space,''
\newblock in {\em Advances in neural information processing systems}, 2017, pp.
  5099--5108.

\bibitem{li2018pointcnn}
Yangyan Li, Rui Bu, Mingchao Sun, Wei Wu, Xinhan Di, and Baoquan Chen,
\newblock ``Pointcnn: Convolution on x-transformed points,''
\newblock in {\em Advances in neural information processing systems}, 2018, pp.
  820--830.

\bibitem{liu2019relation}
Yongcheng Liu, Bin Fan, Shiming Xiang, and Chunhong Pan,
\newblock ``Relation-shape convolutional neural network for point cloud
  analysis,''
\newblock in {\em IEEE Conference on Computer Vision and Pattern Recognition},
  2019, pp. 8895--8904.

\bibitem{bucilua2006model}
Cristian Buciluǎ, Rich Caruana, and Alexandru Niculescu-Mizil,
\newblock ``Model compression,''
\newblock in {\em ACM SIGKDD international conference on Knowledge discovery
  and data mining}, 2006, pp. 535--541.

\bibitem{hinton2015distilling}
Geoffrey Hinton, Oriol Vinyals, and Jeff Dean,
\newblock ``Distilling the knowledge in a neural network,''
\newblock {\em arXiv preprint arXiv:1503.02531}, 2015.

\bibitem{wang2018kdgan}
Xiaojie Wang, Rui Zhang, Yu~Sun, and Jianzhong Qi,
\newblock ``Kdgan: Knowledge distillation with generative adversarial
  networks,''
\newblock in {\em Advances in Neural Information Processing Systems}, 2018, pp.
  775--786.

\bibitem{xu2017training}
Zheng Xu, Yen-Chang Hsu, and Jiawei Huang,
\newblock ``Training shallow and thin networks for acceleration via knowledge
  distillation with conditional adversarial networks,''
\newblock {\em arXiv preprint arXiv:1709.00513}, 2017.

\bibitem{zhang111adversarial}
Haoran Zhang, Zhenzhen Hu, Wei Qin, Mingliang Xu, and Meng Wang,
\newblock ``Adversarial co-distillation learning for image recognition,''
\newblock {\em Pattern Recognition}, vol. 111, pp. 107659.

\bibitem{liu2019exploiting}
Jian Liu, Yubo Chen, and Kang Liu,
\newblock ``Exploiting the ground-truth: An adversarial imitation based
  knowledge distillation approach for event detection,''
\newblock in {\em Proceedings of the AAAI Conference on Artificial
  Intelligence}, 2019, vol.~33, pp. 6754--6761.

\bibitem{tian2019contrastive}
Yonglong Tian, Dilip Krishnan, and Phillip Isola,
\newblock ``Contrastive representation distillation,''
\newblock in {\em International Conference on Learning Representations}, 2019.

\bibitem{scheirer2014probability}
Walter~J Scheirer, Lalit~P Jain, and Terrance~E Boult,
\newblock ``Probability models for open set recognition,''
\newblock {\em IEEE transactions on pattern analysis and machine intelligence},
  vol. 36, no. 11, pp. 2317--2324, 2014.

\bibitem{zhang2020hybrid}
Zhang Hongjie, Li~Ang, Guo Jie, and Guo Yanwen,
\newblock ``Hybrid models for open set recognition,''
\newblock in {\em European Conference on Computer Vision}, 2020, pp.
  1225--1232.

\bibitem{sun2020conditional}
Xin Sun, Zhenning Yang, Chi Zhang, Keck-Voon Ling, and Guohao Peng,
\newblock ``Conditional gaussian distribution learning for open set
  recognition,''
\newblock in {\em IEEE Conference on Computer Vision and Pattern Recognition},
  2020, pp. 13480--13489.

\bibitem{yoshihashi2019classification}
Ryota Yoshihashi, Wen Shao, Rei Kawakami, Shaodi You, Makoto Iida, and Takeshi
  Naemura,
\newblock ``Classification-reconstruction learning for open-set recognition,''
\newblock in {\em IEEE Conference on Computer Vision and Pattern Recognition},
  2019, pp. 4016--4025.

\bibitem{wu20153d}
Zhirong Wu, Shuran Song, Aditya Khosla, Fisher Yu, Linguang Zhang, Xiaoou Tang,
  and Jianxiong Xiao,
\newblock ``3d shapenets: A deep representation for volumetric shapes,''
\newblock in {\em IEEE conference on computer vision and pattern recognition},
  2015, pp. 1912--1920.

\end{thebibliography}

\end{document}